\documentclass[a4paper,fleqn]{cas-dc}

\usepackage[round,authoryear]{natbib}
\usepackage{subfig}
\usepackage{amsmath}
\usepackage[ruled]{algorithm2e}
\usepackage{bm}
\usepackage{hyperref}
\usepackage{color, xcolor}
\usepackage{graphicx}
\usepackage{xpatch}

\def\tsc#1{\csdef{#1}{\textsc{\lowercase{#1}}\xspace}}
\tsc{WGM}
\tsc{QE}

\makeatletter
\def\changeBibColor#1{%
	\in@{#1}{cui2024rakcr, qin2024heterogeneous, qiu2024multi, meng2024deep, shou2025masked, zhang2024groupface, zhou2024facilitating, bao2023general, zhao2024mixture}
	\ifin@\color{black}\else\normalcolor\fi
}

\xpatchcmd\@bibitem
{\item}
{\changeBibColor{#1}\item}

\xpatchcmd\@lbibitem
{\item}
{\changeBibColor{#2}\item}
{}{\fail}
\makeatother

\begin{document}
\let\WriteBookmarks\relax
\def\floatpagepagefraction{1}
\def\textpagefraction{.001}
\let\printorcid\relax

\shortauthors{Yiping~Zhang et~al.}


\title [mode = title]{LRA-GNN: Latent Relation-Aware Graph Neural Network with Initial and Dynamic Residual for Facial Age Estimation}

\author[a]{Yiping Zhang}
\ead{yipingzhang@csuft.edu.cn}

\author[a]{Yuntao Shou}
\ead{shouyuntao@stu.xjtu.edu.cn}

\author[a]{Wei~Ai}
\ead{weiai@csuft.edu.cn}

\author[a]{Tao~Meng}  
\cormark[1]
\ead{mengtao@hnu.edu.cn}
\cortext[cor1]{Corresponding author}

\author[b]{Keqin~Li}
\ead{lik@newpaltz.edu}


\address[a]{College of Computer and Mathematics, Central South University of Forestry and Technology, Changsha, Hunan 410004, China.}
\address[b]{Department of Computer Science, State University of New York, New Paltz, New York 12561, USA.}

\begin{abstract}
Face information is mainly concentrated among facial key points, and frontier research has begun to use graph neural networks to segment faces into patches as nodes to model complex face representations. However, these methods construct node-to-node relations based on similarity thresholds, so there is a problem that some latent relations are missing. These latent relations are crucial for deep semantic representation of face aging. In this novel, we propose a new Latent Relation-Aware Graph Neural Network with Initial and Dynamic Residual (LRA-GNN) to achieve robust and comprehensive facial representation. Specifically, we first construct an initial graph utilizing facial key points as prior knowledge, and then a random walk strategy is employed to the initial graph for obtaining the global structure, both of which together guide the subsequent effective exploration and comprehensive representation. Then LRA-GNN leverages the multi-attention mechanism to capture the latent relations and generates a set of fully connected graphs containing rich facial information and complete structure based on the aforementioned guidance. To avoid over-smoothing issues for deep feature extraction on the fully connected graphs, the deep residual graph convolutional networks are carefully designed, which fuse adaptive initial residuals and dynamic developmental residuals to ensure the consistency and diversity of information. Finally, to improve the estimation accuracy and generalization ability, progressive reinforcement learning is proposed to optimize the ensemble classification regressor. Our proposed framework surpasses the state-of-the-art baselines on several age estimation benchmarks, demonstrating its strength and effectiveness.
\end{abstract}

\maketitle
	




\textcolor{black}{\footnote{Our code is publicly available at \href{https://github.com/Bottle010/LRA-GNN}{https://github.com/Bottle010/LRA-GNN}.}}
		
\section{Introduction}
With the continuous development of deep learning, the field of face recognition shows an increasingly prosperous trend. As one of the important attributes of the face, age is also an important topic in current face research. The potential value of age has been gradually discovered and applied in numerous fields, such as human-computer interaction \citep{yang2018ssr, shou2022conversational, shou2025masked, shou2022object, shou2023comprehensive, shou2024adversarial, meng2024deep}, social media \citep{duan2017ensemble, shou2023adversarial, ai2023gcn, ai2024gcn, meng2024multi, shou2024contrastive}, and video surveillance \citep{rothe2015dex, shou2024spegcl, shou2024efficient, ying2021prediction}.

However, age estimation tasks continue to be challenging due to the complicated internal and external factors \citep{agbo2021deep, zhang2019fine, shou2023graph, shou2023czl}. Early age estimation methods were mainly based on manual feature extractors and machine learning algorithms, which were difficult to deal with the increase in image complexity and data size. With the increase in computational power, image feature extraction based on Convolutional Neural Network (CNN) has shown outstanding performance.  \citet{rothe2015dex, meng2024masked, shou2024revisiting, ai2024edge} utilized a pre-trained VGG-16 network for facial representation and obtained an age regression result through classification probabilities multiplied by the corresponding labels. \citet{zhang2019fine, zhang2024multi, ai2023two} integrated LSTM units with the residual networks (ResNets) to extract local age-sensitive features and used the Deep EXpectation algorithm (DEX) for age regression. Moreover, with the success of Transformer, many efforts have been based on the Vision Transformer (ViT) \citep{dosovitskiy2020image, meng2024revisiting, ai2024mcsff} for long-range dependency modeling.  \citet{kuprashevich2023mivolo, shou2023graphunet, ai2024graph, shou2024low} proposed Multi Input VOLO (MiVOLO) utilizing the newest vision transformer for age and gender estimation in the wild. \citet{qin2023swinface, shou2024graph, ai2024seg, ai2024contrastive} proposed SwinFace based on ViT for multi-task facial feature extraction including age estimation.

Despite the CNN and Transformer methods have outstanding capabilities in image processing, they cannot be applied to data in non-Euclidean spaces. In particular, facial attributes are mainly defined around the specific facial key points, so these modeling approaches are inflexible for complex and irregular human faces. Therefore, remarkable works have utilized Graph Neural Networks (GNN) to model facial structural information by directly learning potential embedded nodes based on their neighbors and relationships. \citet{korban2023taa} proposed a Time-Aware Adaptive Graph Convolutional Network (TAA-GCN) to compute the temporal age dependence using Temporal Memory Module (TMM). \citet{shou2025masked} proposed a Masked Contrastive Graph Representation Learning (MCGRL) to capture the rich structural information of the face, which outperforms the CNN and Transformer based methods. \citet{ge2024mgrr, shou-etal-2025-dynamic} proposed facial action units detection network MGRR-Net with multi-level graph feature learning and relational reasoning.

\begin{figure}
	\centering
	\color{black}
	\includegraphics[width=1\linewidth]{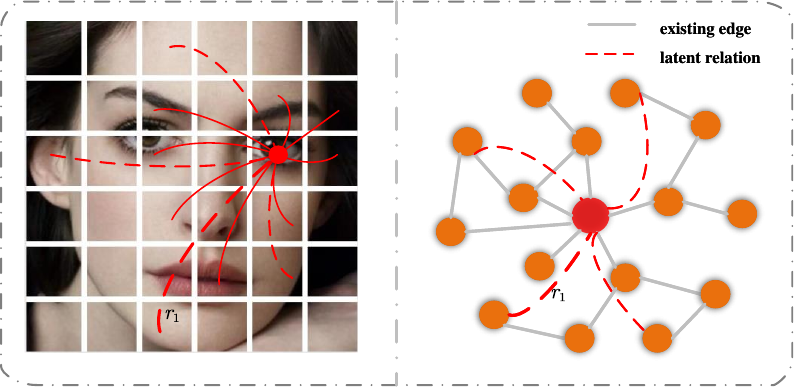}
	\caption{The illustration of the latent relation (e.g., the dashed part). Existing GNN-based methods utilize the similarity threshold method and may ignore relations between facial key points and wrinkles such as $r_1$, which are not similar enough but important to attain more accurate estimation.}
	\label{fig_1}
\end{figure}

All of the above methods have achieved excellent performance, but they still have some deficiencies. (1) \textbf{Insufficient in capturing latent relations.} GNN-based methods consider explicit modeling relationships based on similarity thresholding, but they cannot capture the latent relations of facial key points. The latent relations are schematically shown in Figure \ref{fig_1}, which are indispensable for the deep semantic representation of face aging. Recent work has begun to focus on this point, \citet{jiang2023face2nodes} utilizes the dilated K-nearest neighbors algorithm to learn the latent connections between face graph nodes. However, this approach is limited and incomplete for capturing latent relations. (2) \textbf{Inadequate to co-optimization of multi-stage age estimation.} Existing age estimation often utilizes hybrid algorithms, but the optimization of different stages is learned independently. \citet{duan2017ensemble} introduced an extreme learning machine (ELM) to optimize age estimation, but the ELM classifier and ELM regressor were trained separately, which is easy to fall into the suboptimal problem.

To settle the two problems above together, we propose a new Latent Relation-Aware Graph Neural Network with Initial and Dynamic Residual (LRA-GNN) for robust and comprehensive facial representation, as well as design the progressive reinforcement learning to synergistically optimize multi-stage age estimation. Specifically, face images are segmented into patches of the same size as graph nodes, and we utilize facial key points as prior knowledge to construct an initial graph. Then we employ a random walk strategy on the initial graph to obtain the global structure, which can collect more information with fewer paths in the search space. Under the guidance of the aforementioned strategies, LRA-GNN leverages the multi-attention mechanism to capture the latent relations and generates a set of fully connected graphs containing rich facial information and complete structure. Unlike the commonly used similarity computing \citep{korban2023taa,jang2024multi} and K-nearest neighbor algorithms \citep{shou2025masked,jiang2023face2nodes}, our mechanism can cover all relations without omission. In addition, to avoid over-smoothing issues for deep feature extraction on the fully connected graphs, we design the deep residual graph convolution networks to convolve the fully connected graph. These networks fuse adaptive initial residuals and dynamic developmental residuals to ensure the consistency and diversity of information, which are powerful for deep representation learning.

After fed the robust and comprehensive facial features into age estimation, to improve the accuracy and generalization of our model, we propose the progressive reinforcement learning to synergistically optimize the subsequent age group classification and final regression. We define age estimation by classification and then regression as a walk-to-the-end problem on a grid, which emphasize the continuity and correlation of ages unlike common categorical regression. In addition, to bootstrap agents to be generalizable for age estimation, we carefully design reward functions which takes into account the age distribution of different age groups for more robust age grouping and more accurate regression prediction. When the agent possesses the optimum behavior to maximize the cumulative reward through interaction with the environment, it can perform step-wise age estimation by classification and then regress the samples as correctly as possible.

\subsection{Our Contributions}
Extensive experiments have demonstrated that the model we proposed is flexible, effective, and comprehensive. Our contributions to this paper are summarized as follows:

\begin{itemize}
	\item We propose a new Latent Relation-Aware Graph Neural Network with Initial and Dynamic Residual (LRA-GNN) for obtaining robust and comprehensive facial representations.
	
	\item To effectively explore the semantic and structural information of faces, we utilize the face key points as prior knowledge to construct an initial graph and apply a random walk strategy to the initial graph to obtain global structure.
	
	\item We carefully design a multi-attention mechanism to capture latent relations and a deep residual graph convolution network that fuses adaptive initial residuals and dynamic developmental residuals to ensure the consistency and diversity of information.
	
	\item To improve the generalization ability of our architecture, we propose progressive reinforcement learning to synergistically optimize the age group classification and final regression. The robust reward function and loss function are designed together to guide our age estimation.
	
\end{itemize}

\section{Related work}
\label{sec:sec2}
In this section, we first introduce the frontier research of graph neural networks. Then we review some related works in deep learning-based age estimation. Finally, we briefly describe the reinforcement learning.

\subsection{Graph Neural Networks}
Graph Neural Networks (GNNs) have achieved impressive results in the field of processing non-Euclidean data and become a popular research. \textcolor{black}{
They have been widely applied in several tasks such as recommendation systems \citep{cui2024rakcr}, image retrieval \citep{qin2024heterogeneous}, action recognition \citep{qiu2024multi}, and multi-modal emotion recognition \citep{meng2024deep}.
}

There are several variants developed from GNNs, such as Graph Convolutional Neural Networks (GCNs) \citep{kipf2016semi}, Graph Attention Networks (GATs) \citep{velivckovic2017graph}, GraphSAGE \citep{hamilton2017inductive} and so on. For the most commonly used GCNs, existing work mainly focuses on two streams: spectral-based \citep{henaff2015deep} and spatial-based \citep{atwood2016diffusion, niepert2016learning}. However, they utilized the shallow networks that limit their representation capabilities, while deep GCNs are prone to over-smoothing which results in a rapid decrease in node differentiation. To get over this weakness, the mainstream attempts are two-fold:  \citet{li2019deepgcns} proposed DeepGCN similar to DeepCNN utilizing dilated convolutions, dense or residual connections to improve the expressive power. \citet{abu2019mixhop} proposed MixHop which involved discerning neighborhood connections at different distances by iteratively mixing the feature embedding. 

\textcolor{black}{
Following the former, we integrate a multi-head attention mechanism with a deep residual graph convolutional network fusing adaptive initial residuals and dynamic developmental residuals to obtain robust and comprehensive facial representation.
}

\subsection{Age Estimation}
Age estimation stands as a crucial and difficult task within the realm of computer vision. 
\textcolor{black}{
Current age estimation methods usually work on designing more robust face representation networks \citep{kuprashevich2023mivolo, shou2025masked, zhang2024groupface} or more efficient age estimation techniques \citep{shin2022moving, wang2023exploiting, chen2023daa}.
}

\textcolor{black}{
\citet{shin2022moving} proposed moving window regression (MWR), which formed a search window through two reference instances and then estimated the rho-rank. \citet{chen2023daa} proposed the Delta Age AdaIN (DAA), utilizing binary code mapping and age encoder-decoder.  \citet{kuprashevich2023mivolo} proposed Multi Input VOLO (MiVOLO) utilizing the newest vision transformer for age and gender estimation in the wild. \citet{wang2023exploiting} utilized meta-learning paradigm to built an unfair filtering network that reduce category bias in age estimation. \citet{zhang2024groupface} proposed GroupFace, integrating a multi-hop attention GCN with a group-aware margin strategy, which is effective in imbalanced age estimation. \citet{shou2025masked} proposed a Masked Contrastive Graph Representation Learning (MCGRL) to capture the rich structural information of the face, which outperforms the CNN and Transformer based methods. However, these methods is sufficient in capturing latent relations.
}

\textcolor{black}{
In this novel, we design the multi-head attention mechanism with a deep GCN to capturing latent relations effectively and efficiently.}

\subsection{Reinforcement Learning}
Reinforcement learning (RL) has become a new paradigm in artificial intelligence technology, which has also gained high speed in recent years. RL has shown a broad application prospect in the fields of finance, gaming, automation, robotics, and so on. The RL aims to train intelligence to make autonomous decisions by maximizing future cumulative rewards, which can effectively optimize image classification and regression. \citet{lin2020deep} considered imbalanced data classification as a Markov decision process, where samples are states, classification is actions, and rewards are based on the match between predicted and true values. \citet{wen2021building} modeled the decision tree generation as a Markov decision process and achieved significant results by guiding tree construction through reinforcement learning. \citet{yang2023deep} further optimized the original reinforcement learning algorithm to avoid overestimation of values and improve the training effectiveness. 

In this novel, the Double Deep Q Network (DDQN) is introduced to reinforcement learning for achieving better classification and prediction results, which identifies high-level features using a reward function that distinguishes between different classes, i.e., punish minorities more harshly or reward them more generously.

\section{Methodology}
\label{sec:sec3}
\begin{figure*}[!t]
	\centering
	{\includegraphics[width=1\textwidth]{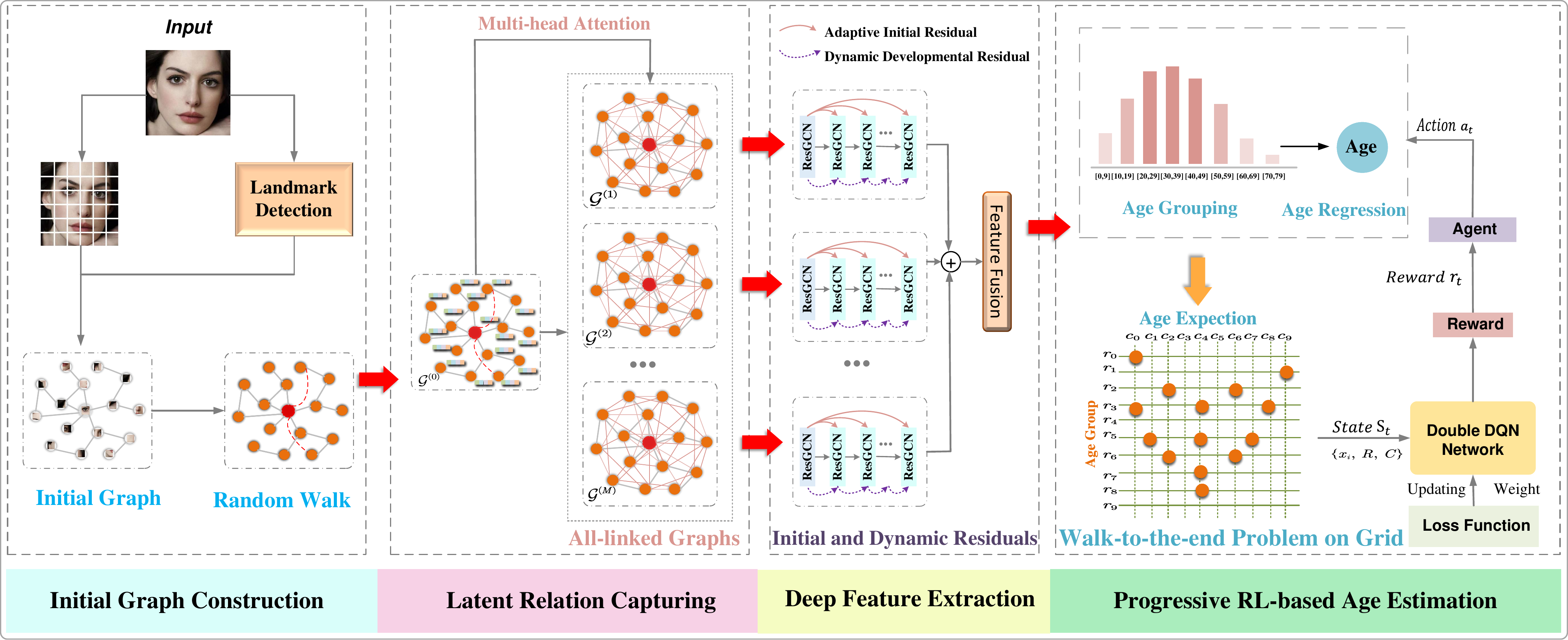}}
	\caption{\textbf{The Overall Pipeline} of our Latent Relation-Aware Graph Neural Network with Initial and Dynamic Residual (\textbf{LRA-GNN}). First, utilizing facial key points as prior knowledge, face images are segmented into patches as graph nodes to construct an initial graph. Then LRA-GNN leverages the multi-attention mechanism to capture the latent relations and generates a set of fully connected graphs. Finally, this set of fully connected graphs is fed into deep residual graph convolutional networks for feature extraction and through progressive reinforcement learning to achieve robust and accurate age estimation.}
	\label{fig_2}
\end{figure*}

In this section, we are set to furnish an exhaustive delineation of our proposed approach the Latent Relation-Aware Graph Neural Network with Initial and Dynamic Residual (LRA-GNN) for feature extraction and the Progressive Reinforcement Learning-based Age Estimation. The overall pipeline is shown in Figure \ref{fig_2}.

\subsection{Overview of Our Framework}
The main design of our framework is organized around the following questions:

\textbf{Q1:} \emph{What should pay attention to get robust and comprehensive facial representation?}

\textbf{Q2:} \emph{How to capture latent relations effectively and efficiently?}

\textbf{Q3:} \emph{How to extract deep features based on the obtained fully connected graphs?}

\textbf{Q4:} \emph{What are the strategies to synergistically optimize age estimation for higher accuracy and better generalization?}

In feature extraction, the face image is first segmented into equal-sized patches as graph nodes. \textbf{Q1:} As shown in Algorithm 1, to get robust and comprehensive facial representation, we propose a new Latent Relation-Aware Graph Neural Network with Initial and Dynamic Residual (LRA-GNN) for capturing the latent relations. \textbf{Q2:} Specifically, we first construct an initial graph utilizing facial key points as prior knowledge, and then a random walk strategy is employed to the initial graph for obtaining the global structure, both of which together guide the subsequent effective exploration and comprehensive representation. Then LRA-GNN leverages the multi-attention mechanism to capture the latent relations and generates a set of fully connected graphs containing rich facial information and complete structure based on the aforementioned guidance. \textbf{Q3:} In addition, to avoid over-smoothing issues for deep feature extraction on the fully connected graphs, we design the deep residual graph convolution networks to convolve the fully connected graph. These networks fuse adaptive initial residuals and dynamic developmental residuals to ensure the consistency and diversity of information, which are powerful for deep representation learning.

After feeding the comprehensive and robust facial features into age estimation, \textbf{Q4:} to improve the generalization and accuracy of our framework, we propose progressive reinforcement learning to optimize the subsequent age group classification and final regression synergistically. We define age estimation by classification and then regression as a walk-to-the-end problem on a grid. With a well-designed reward function, a trained agent can perform step-wise age estimation as correctly as possible.

\subsection{Initial Graph Construction}
\subsubsection{Graph Construction}
Face attributes are mainly concentrated at facial key points, and we first employ a facial landmark algorithm to select the most information-rich representations. Existing face landmark detection algorithms can easily give the $2D$ coordinates of the facial key points, and we utilize the algorithm \citep{korban2023taa} to select the key points that are susceptible to age changes and insensitive to facial expressions. Each face image has $n$ keypoints as $U_i=\left\{ u_1,\,\,u_2,\,\,...\,\,,\,\,u_n \right\} $, each keypoint contains coordinate information $u_i=\left( x,y \right) $.

Then the original face image can be partitioned into multiple same-size patches, where the similar pixel blocks around the keypoints will be stitched in the same patch as much as possible based on the coordinate information.  Formally, assume that the input image $I$ has shape $H\times W\times 3$, which is segmented to $N$ patches corresponding to the key points, and the patches are embedded into $M$ feature dimensions forming a feature vector. Each latent vector is treated as a node $\mathcal{V}=\left\{ v_1,\,\,v_2,\,\,...\,\,,\,\,v_N \right\} $, yielding us a graph representation $
\mathcal{G}=\left(\mathcal{V},\,\,\mathcal{E},\,\,\mathcal{A} \right) $, where $E$ is the edges set, $A$ denotes the adjacency matrix with initializing to a 0-1 matrix by the similarity computing method. Moreover, let $\mathcal{X}\in \mathbb{R}^{N\times M}$ be the node features and $\mathcal{E}\in \mathbb{R}^{N\times M}$ be the edge features, where $x_i$ denotes the node embedding and $e_{ij}$ denotes the edge embedding.

\subsubsection{Random Walk Updating}
\textcolor{black}{
The key to graph representation learning is the aggregation and propagation of domain node information, which is closely related to paths.
Therefore, to initially obtain the global structure, the random walk strategy \citep{grover2016node2vec, zhou2024facilitating} is employed in the initial graph to collect more information with fewer paths in the search space.
}

Given a random walk that uniformly samples a random vertex $v_i$, which has visited edge $e_{ik}$ to reach node $v_k$, assume that the next walk will visit $v_j$, and the non-normalized transfer probability formula for the walk is:
\begin{equation}
		\mathcal{P}\left( v_j|\,\,v_k,\,\,v_i \right) =\left\{ \begin{array}{l}
			\,\,1/p,\,\,\,\,if\,\,d_{ij}=0\\
			\,\,1,\,\,\,\,\,\,\,\,\,\,if\,\,d_{ij}=1\\
			\,\,1/q,\,\,\,\,if\,\,d_{ij}=2\\
			\,\,0,\,\,\,\,\,\,\,\,otherwise\\
		\end{array} \right.
	\label{equ1}
\end{equation}
where $m$, $n$ are the adjustable parameters and $d_{ij}$ is the shortest path length of node $v_i$ to $v_j$. The random walk employs Depth First Search (DFS) when $p>1$ and $q<1$ while employing Breadth First Search (BFS) when $p<1$ and $q>1$.

By optimizing the problem of random walk, it updates the adjacency matrix by different walk paths. Calculating the cosine similarity of higher-order structural information can determine whether there are stronger connections between nodes in the global structure and the update formula can be expressed as:

\begin{equation}
	\begin{aligned}
		&\tilde{A}_{ij}=A_{ij}+f\left( i,j \right) ,
		\\
		&f\left( i,j \right) =\left\{ \begin{array}{l}
			1,\ if\ \cos \left( x_i,x_j \right) \ge \tau\\
			0,\ otherwise\\
		\end{array} \right.
	\end{aligned}
	\label{equ2}
\end{equation}
where $\cos \left( \cdot \right) $ denotes the cosine similarity calculation and $\tau$ is the threshold different from initial similarity.

\begin{figure}[!t]
	\centering
	\includegraphics[width=0.48\textwidth]{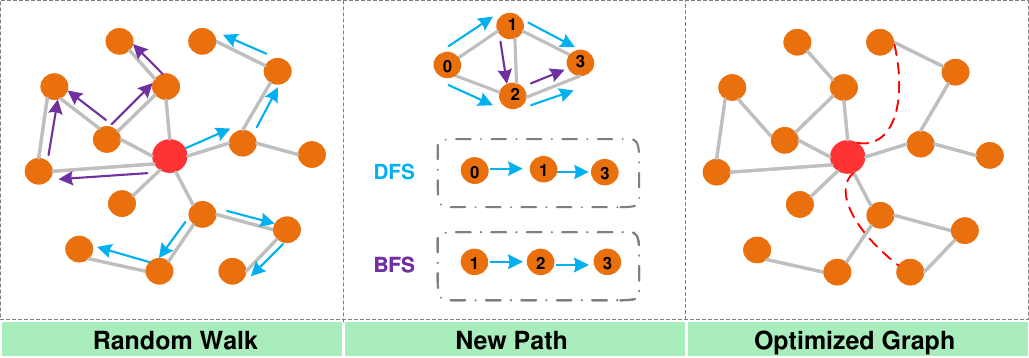}
	\caption{The illustration of Random Walk Updating. BFS tends to visit the immediate neighbors of the source node, and DFS tends to explore nodes that are further and further away from the source node. The combination of the two can effectively capture the local and global relations between nodes. }
	\label{fig_3}
\end{figure}

\textcolor{black}{
As shown in Figure \ref{fig_3}, after utilizing the random walk strategy, we update the graph representation and obtain a new optimized graph embedding $\tilde{\mathcal{G}}=\left( \tilde{\mathcal{V}},\,\,\tilde{\mathcal{E}},\,\,\tilde{\mathcal{A}} \right)$.
With different similarity thresholds, the updated graph embeds a few more explicit relations.
This not only initially captures the global structural features in the graph, but also significantly reduces the number of nodes and edges that need to be taken into account, reducing the burden of the subsequent graph convolution and speeding up the computational process.
}

\subsection{Latent Relation Capturing}
\textcolor{black}{
After explicit exploration of semantic and structural facial information using random walk, to obtain more comprehensive facial representation, the multi-head attention \citep{ren2021lr} is then applied on the optimized initial graph embedding $\tilde{\mathcal{G}}$  to explore latent relations between facial keypoints.
} Specifically, the multi-head self-attention is performed to construct a set of fully connected graphs that integrates the correlations between vertices. Generating m-indexed fully connected graphs can be formulated as:

\begin{equation}
		\tilde{A}^{\left( m \right)}=soft\max \left( \frac{\tilde{X}_iW_{m}^{i}\times \left( \tilde{X}_jW_{M}^{j} \right) ^T}{\sqrt{M}} \right) \tilde{A}^{\left( 0 \right)}
	\label{equ3}
\end{equation}
where $\tilde{A}^{\left( 0 \right)}$ is the initial adjacency matrix of optimized graph $\tilde{\mathcal{G}}$, $\tilde{X}_i$ and $\tilde{X}_j$ denotes the node embedding from the node set $\tilde{\mathcal{V}}=\left\{ v_1,\,\,...\,\,,\,\,v_i,\,\,...\,\,,\,\,v_j,\,\,...\,\,v_N \right\} $. $M$ is feature vector’s dimension, while $W_{m}^{i}$ and $W_{m}^{j}$ are the pairwise transfer matrices for $\tilde{X}_i$ and $\tilde{X}_j$, respectively. From this mechanism, aggregation by fully connected graphs can capture latent relations between facial key points more comprehensively, increase the diversity of correlations, and address the weakness of common GNNs missing some important information.

\subsection{Deep Feature Extraction}
To extract deep and rich facial feature information, we perform the graph convolution operation on the obtained $m$ different fully connected graphs.

DeepGCNs can expand the receptive field and improve the graph model representation performance, which can be formulated as:
\begin{equation}
		H^{\left( l+1 \right)}=\text{Re}LU\left( \tilde{D}^{-\frac{1}{2}}\tilde{A}\tilde{D}^{-\frac{1}{2}}H^{\left( l \right)}W^{\left( l \right)} \right)
	\label{equ4}
\end{equation}
where $\tilde{D}=\sum_j{\tilde{A}_{ij}}$ denotes the degree matrix, $\tilde{A}=A+I$, $I$ is the unit matrix, $H^{\left( l \right)}$ is the $l$-th layer embedding of the graph nodes and $W^{\left( l \right)}$ is the learnable weight.

\textcolor{black}{
However, GCNs stack too many layers tending to over-smoothing, where the representations of the nodes converge and become indistinguishable. For the fully connected graphs generated by multi-head attention mechanisms, the over-smoothing problem caused by deepening the number of layers of GCNs is more severe and needs to be urgently addressed. Some works have begun to introduce residual connections to alleviate the over-smoothing problem and it's working well}, ResGCN \citep{li2019deepgcns} is expressed as: 
\begin{equation}
	\resizebox{.9\hsize}{!}{
		$H^{\left( l+1 \right)}=\text{Re}LU\left( \left( 1-\alpha \right) \tilde{D}^{-\frac{1}{2}}\tilde{A}\tilde{D}^{-\frac{1}{2}}H^{\left( l \right)}W^{\left( l \right)}+\alpha H^{\left( l-1 \right)} \right)  $ }
	\label{equ5}
\end{equation}
where $W^{\left( l \right)}$ is the residual connection weight parameter. And the initial residual connection replaces the node representation of the previous layer by combining $H^{\left( l-1 \right)}$ with the initial representation $H^{\left( 0 \right)}$.

\begin{figure}[!t]
	\centering
	\includegraphics[width=0.48\textwidth]{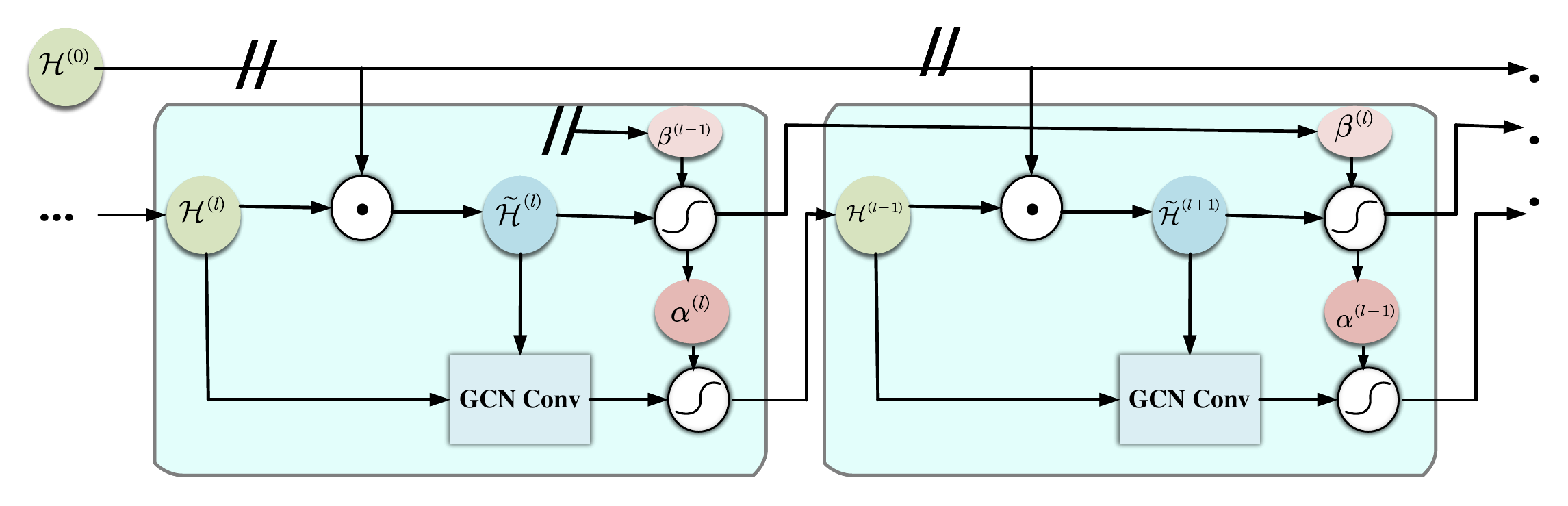}
	\caption{
	The illustration of deep feature extraction with the adaptive initial residuals and dynamic developmental residuals. \textcolor{black}{The adaptive initial residuals obtain the personalized characteristics from initial embedding $H^{\left( 0 \right)}$ and hidden embedding $H^{\left( l \right)}$. The dynamic developmental residuals gain the developmental pattern $\alpha ^{\left( l \right)}$ from the residual embedding $  \tilde{H}^{\left( l \right)}$.}
	}
	\label{fig_4}
\end{figure}

\textcolor{black}{Nevertheless, ResGCN uses fixed values to represent layer-to-layer node connections, ignoring the individualized characteristics of the initial nodes and the dynamically developmental correlations between layers. And learning the personalized characteristics of initial nodes and the related properties of dynamic development between layers can not only solve the over-smoothing problem, but also enhance the expressive power of GCNs.
}
Inspired by the \citep{zhang2023drgcn}, we design the deep residual GCNs to combine the adaptive initial residuals and dynamic developmental residuals. The adaptive initial residuals can adaptively derive information from the initial representation to prevent noise-mitigating over-smoothing problems, which can be formulated as:
\begin{equation}
		\tilde{H}^{\left( l+1 \right)}=\left( 1-\beta ^{\left( l \right)} \right) \tilde{D}^{-\frac{1}{2}}\tilde{A}\tilde{D}^{-\frac{1}{2}}H^{\left( l \right)}+\beta ^{\left( l \right)}H^{\left( 0 \right)}
	\label{equ6}
\end{equation}
where $\beta ^{\left( l \right)}$ denotes the proportion of adaptively connected initial features at layer $l$.

The dynamic developmental residuals capture the residual developmental state of the layers and avoid stacking too many non-linear mappings resulting in vanishing gradients, which can be formulated as:
\begin{equation}
		\resizebox{.9\hsize}{!}{$
			H^{\left( l+1 \right)}=\text{Re}LU\left( \left( 1-\alpha ^{\left( l \right)} \right) \tilde{D}^{-\frac{1}{2}}\tilde{A}\tilde{D}^{-\frac{1}{2}}\tilde{H}^{\left( l \right)}W^{\left( l \right)}+\alpha ^{\left( l \right)}\tilde{H}^{\left( l-1 \right)} \right)  $}
	\label{equ7}
\end{equation}
where $\alpha ^{\left( l \right)}=\varPsi \left( \tilde{H}^{\left( l \right)},\,\,\tilde{H}^{\left( l-1 \right)} \right)$ denotes the dynamic developmental factor to adjust the percentage of information retained from the previous layer, $\varPsi \left( \cdot \right) $ is the developmental function.

\begin{table}[htbp]
	\centering
	\renewcommand\arraystretch{1.5}
	\setlength{\tabcolsep}{1pt}{
		\resizebox{\linewidth}{!}{
			\begin{tabular}{l}
				\toprule
				\textbf{Algorithm 1} Latent Relation-Aware Graph Neural Network \\ with Initial and Dynamic Residual (LRA-GNN).
				\\ \hline
				\textbf{Input:} Images $\mathcal{I}=\left\{ I_1,\,\,I_2,\,\,...\,\,,\,\,I_n \right\} $, multi-head number $M$;
				\\
				\textbf{Output:} Comprehensive facial feature embedding  $\mathcal{H}$.  \\
				1: \textbf{for} number of training epochs \textbf{do}\\
				2:  \qquad  \textbf{for} ${I_i}$ \textbf{in} all images  \textbf{do} \\
				\qquad \qquad \quad  \textbf{// Initial Graph Construction} \\
				3: \qquad \qquad Construct an initial graph $\mathcal{G}=\left( \mathcal{V},\,\,\mathcal{E},\,\,\mathcal{A} \right)$ \\ \qquad \qquad \quad  using facial keypoints as priori knowledge. \\
				\qquad \qquad \quad  \textbf{// Random Walk Updating} \\
				4: \qquad \qquad Obtain $\tilde{\mathcal{G}}=\left( \tilde{\mathcal{V}},\,\,\tilde{\mathcal{E}},\,\,\tilde{\mathcal{A}} \right) $ on initial graph by Eq. (1), (2). \\
				\qquad \qquad \quad  \textbf{// Latent Relation Capturing} \\
				5: \qquad \qquad \textbf{for} ${v_i}$ \textbf{in} $\tilde{\mathcal{V}}$  \textbf{do}   \\
				6: \qquad \qquad \qquad Generate fully connected graphs by Eq. (3). \\
				7: \qquad \qquad \textbf{end for} \\
				\qquad \qquad \quad  \textbf{// Deep Feature Extraction} \\
				8:\qquad \qquad  \textbf{for} ${v_i}$ \textbf{in} $\tilde{\mathcal{V}}$  \textbf{do} \\
				9: \qquad \qquad \qquad Calculate adaptive initial residual by Eq. (6). \\
				10:\qquad \qquad \qquad Calculate dynamic developmental residual by Eq. (7).\\
				11:\qquad \qquad \qquad Fusing two residuals to obtain embedding $\mathcal{H}$.\\
				12: \qquad \qquad \textbf{end for} \\
				13: \qquad \textbf{end for} \\
				14: \textbf{end for} \\
				15: \textbf{Return:} The feature embedding $\mathcal{H}$.\\
				\bottomrule
		\end{tabular}}
	}
	\label{algo1}
\end{table}

\subsection{Progressive RL-based Age Estimation}
\begin{figure}[!t]
	\centering
	\includegraphics[width=0.48\textwidth]{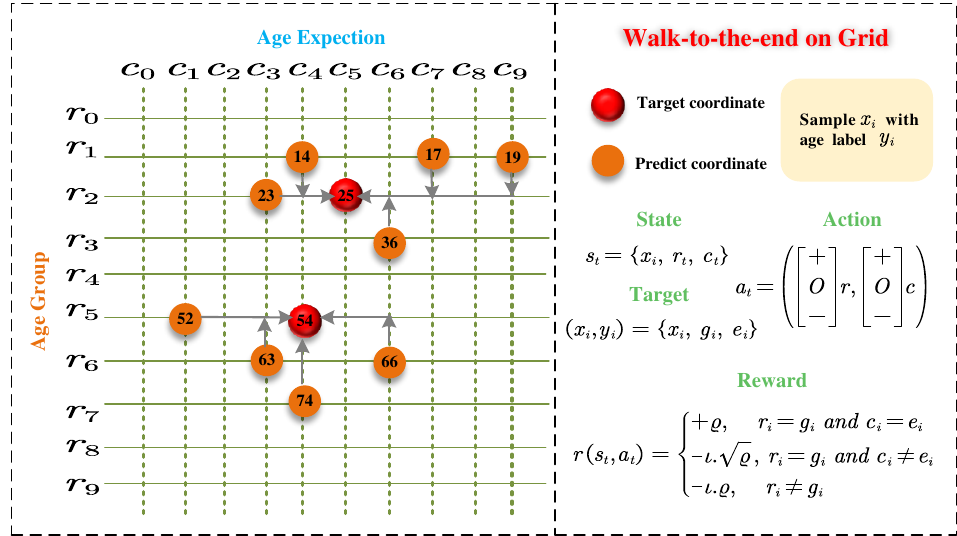}
	\caption{The illustration of Progressive RL-based Age Estimation. The age estimation through classification and then regression is defined as a walk-to-the-end problem on a grid.}
	\label{fig_5}
\end{figure}

We define the age estimation through classification and then regression as a walk-to-the-end problem on a grid. As shown in Figure \ref{fig_5}, the grid is divided into rows and columns, with the rows representing the classified age groups and the columns representing the serialized age regression values within that age group. At each time interval, the agent is presented with a sample and places (classifies) it into the appropriate row (age group), subsequently positioning (regresses) it within one of the columns of that row (predicted value).  Depending on the agent's different actions, the environment provides an immediate reward and the subsequent sample. When the agent moves the sample to the correct row and column, the environment will give the agent a positive reward, otherwise, it will give the agent a penalty. As the agent acquires the optimal conduct through its engagement with the environment to achieve the highest cumulative reward, it can make progressive age estimation through classification and then regression on the samples as correctly as possible.

We model the progressive RL-based age estimation as a Markov Decision Process (MDP) and describe it using a five-tuple $\left\{ S, A, R, P,\varUpsilon \right\} $. In the framework, the state space $s_t\in S$, action space $a_t\in A$, rewards $r_t\in R$, policy $\pi \in P$ and discount factor $\gamma \in \left( 0,1 \right)$.

\subsubsection{State}	
The state of the environment is determined by the trained age estimation samples, which can be denoted as $\left\{ x_i,\,\, R,\,\, C \right\} $. Specifically, the sample $x_i$ with real label $y_i$ can be split into corresponding age groups $g_i\in G$ (rows $r_i\in R$) and values within groups $e_i\in E$ (columns $c_i\in C$). $G$ and $E$ contain 0-9, e.g. $g_1$ denotes the age group 10-19 years, while $e_1$ in $g_1$ denotes the second age value of this age group i.e. 11 years.		

\subsubsection{Action}
Agent's actions are associated with the labels of the training datasets, which are targeted to move to the correct coordinates. The main actions include: row move $d_r$ units, column move $d_c$ units, or none of the row or column move. Represent the coordinates separately by binary groups i.e. $a_t=\left( +d_r, O \right) $ means the row increases by $d_r$ units and the column stays the same, $a_t=\left( O,-d_c \right) $ means row stay the same and column decreases by $d_c$ units.

\subsubsection{Reward}
Rewards are designed based on the engagement between the environment and the agent, which can measure the agent's performance after acting. Because the sample is categorized into the correct age group is a critical step, the reward given for whether the agent chooses the correct row is higher than for choosing the correct column. We also consider the continuity of age distribution to add the distance from the label value on the reward function. Meanwhile, to make age estimation more robust and generalizable, we design the imbalance ratio to guide the agent to more effectively learn actions within an unbalanced dataset. To enhance the identification of samples from the minority class, the algorithm is attuned to the minority class, delivering a greater reward or penalty upon encountering such a sample. The reward function is delineated as follows:

\begin{equation}
		r\left( s_t,a_t \right) =\left\{ \begin{array}{l}
			+\varrho ,\,\,\,\,\,\,\,\,r_i=g_i\,\,and\,\,c_i=e_i\\
			-\iota .\sqrt{\varrho},\,\,r_i=g_i\,\,and\,\,c_i\ne e_i\\
			-\iota .\varrho ,\,\,\,\,\,\,\,\,r_i\ne g_i\\
		\end{array} \right.
	\label{equ8}
\end{equation}
where $\iota =|r_i-g_i|+|c_i-e_i|$ denotes the distance from label value and  $\varrho =\frac{N_{g_M}}{N_{g_i}}$ denotes the imbalance ratio. $N_{g_M}$ is the amount of majority group and $N_{g_i}$ the amount of $i$-th group.

\subsubsection{Policy}
The policy $\pi$ is a mapping that maps states to actions, denoted by $\pi \left( a\ |\ s \right)$. Depending on the currently observed state $s$, the $\pi _{\theta}$ will decide which action $a$ the agent should perform. We employ a random policy to provide a probability distribution for choosing each action in state $s$. This randomness increases the exploratory ability of the agent and prevents it from falling into a local optimal solution. We define the age estimation problem as the strategy $\pi ^*$ that finds the optimal row and column coordinates to maximize the cumulative reward, and the strategy $\pi _{\theta}$ can be viewed as an ensemble classifier regressor with parameter $\varOmega$.

\subsubsection{Deep Q-Learning}
Q-learning is a reinforcement learning approach rooted in value iteration, which pursues the optimum function $\pi^*$. This signifies that, by opting for the most favorable action $a$ within the present state $s$, the agent is positioned to secure the greatest cumulative reward.

The mapping of the action-value function adheres to the $Bellman$'s equation, which can be articulated as:
\begin{equation}
		Q^*\left( s,a \right) =E_{\pi}\left( r_t+\gamma Q\left( s_{t+1},\,\,a_{t+1} \right) |s_t=s,\,\,a_t=a \right)
	\label{equ9}
\end{equation}
where $\gamma \in \left( 0,1 \right)$ falls within the open interval \((0,1)\), embodying the discount factor that mirrors the influence of the current action's impact on the reward value. The action with the largest $Q$ in state $s$ is chosen by the optimal policy $\pi _*\left( s,\ a \right) $, i.e., the greedy policy, which can be formulated as:

\begin{equation}
		\pi ^*\left( s,a \right) =arg\max _aQ\left( s,a \right)
	\label{equ10}
\end{equation}

Since the Deep Q Network selects a greedy strategy for both action selection and action evaluation and uses the same neural network parameters, it is easy to cause an overestimation of the value function during the learning process, i.e., the projected value function exceeds the factual value, which will affect the final optimal policy. To address this problem,  \citet{van2016deep} proposed the Double Q-learning algorithm, which decouples the selection of actions and action evaluation to solve the overestimation problem towards the traditional Deep Q Network (DQN).

The Double-DQN method initially identifies the action that corresponds to the highest $Q$ value within the online network $Q^{\theta}$. Subsequently, it determines the target $Q$ value stemming from the chosen action $a$ in the target network $Q^{\tau}$, which can be formulated as:
\begin{equation}
		\resizebox{.9\hsize}{!}{$
		Y_{t}^{Double\,\,Q}=r_t+\gamma Q\left( s_{t+1},\,\,arg\max _aQ^{\theta}\left( s_{t+1},\,\,a;\,\,\theta _o \right) ;\theta _t \right)  $}
	\label{equ11}
\end{equation}
where $Y_{t}^{Double\,\,Q}$ is the target $Q$ value, $\gamma$ is the discount factor, $\theta _o$ is the online weight and $\theta _t$ is the target weight.

To ensure the Q-value prediction accuracy of the model, we introduce an $\varepsilon $-greedy strategy to guide the action selection process. This strategy selects the best action based on the current Q-value in most cases, i.e., with probability $1-\varepsilon $, to fully utilize the known information. However, to prevent the model from falling into a local optimal solution and to encourage it to explore the unknown state, we simultaneously retain a certain degree of randomness, i.e., randomly selecting actions with a probability of $\varepsilon $. This strategy of balancing exploration and utilization helps the model find the optimal balance between exploring new knowledge and utilizing existing knowledge. During the training process, we make full use of the objective function as well as the empirical playback mechanism, to optimize the performance of the Q-network.

\begin{table}[htbp]
	\centering
	\renewcommand\arraystretch{1.5}
	\setlength{\tabcolsep}{1pt}{
		\resizebox{\linewidth}{!}{
			\begin{tabular}{l}
				\toprule
				\textbf{Algorithm 2} Progressive RL-based Age Estimation(PRLAE).
				\\ \hline
				\textbf{Input:} Training data $D=\left\{ \left( x_1,y_1 \right) ,\,\,\left( x_2,y_2 \right) ,\,\,...\,\,,\,\,\left( x_n,y_n \right) \right\}
				$, \\episode number $K$.
				\\
				\textbf{Output:} Optimized ensemble classification regressor with parameters $\varOmega $.  \\
				1: Initialize simulation environment $E$. \\
				2: \textbf{for} episode $k=1$ \textbf{to} $K$ \textbf{do} \\
				3: \qquad Shuffle the training data $D$. \\
				4: \qquad Initialize the current state $s_t=\left\{ x_t,\,\,r_t,\,\,c_t \right\}$. \\
				5: \qquad Calculate the target coordinate $\left( g_i,e_i \right)$. \\
				6: \qquad \textbf{for} time $t=1$ \textbf{to} $T$ \textbf{do} \\
				7: \qquad \qquad Choose an action based on $\varepsilon $-greedy policy: $a_t=\pi _{\theta}\left( a_t \right)$. \\
				8: \qquad \qquad Calculate the reward $r\left( s_t,a_t \right)$ by Eq (8). \\
				9: \qquad \textbf{end for} \\
				10: \qquad Obtain next state $s_{t+1}=\left\{ x_{t+1},\,\,r_{t+1},\,\,c_{t+1} \right\}$.\\
				11: \qquad  Calculate the target Q-value by Eq. (11). \\
				12: \qquad Update the parameter $\varOmega$ by computing the loss with Eq. (12). \\
				13:  \qquad \textbf{if} reach the maximum cumulative reward \textbf{then} break \\
				14: \textbf{end for} \\
				15: \textbf{Return:} The parameter $\varOmega$.\\
				\bottomrule
		\end{tabular}}
	}
	\label{algo2}
\end{table}

\subsubsection{Training Network}
As shown in Algorithm 2, taking the face sample features as inputs to train Double DQN, the agent traverses all states by moving on the grid with progressive iterative updates while optimizing the ensemble classifier regressor. Inspired by \citep{lin2017focal}, our progressive age estimation loss function consists of focal loss (FL) $\mathcal{L}_{FL}=-\left( 1-p_i \right) ^{\tau}\log \left( p_i \right) $ and average absolute loss (MAE), which can be  formulated as:
\begin{equation}
		\mathcal{L}_{PRLAE}=\eta \mathcal{L}_{FL}+\left( 1-\eta \right) \mathcal{L}_{MAE}
	\label{equ12}
\end{equation}
where $p_i$ represents the model's predicted probability and $\tau $ is an adjustable focusing parameter. Higher values of $\tau $ reduce the loss of easy samples, which allows the model to turn its attention to hard samples. When $\tau =0$, it becomes the standard cross-entropy loss. We set  $\tau =1.3$ in our study for better classification results.

The loss function $\mathcal{L}_{PRLAE}$ can dynamically consider both hard samples and is optimized together with the reward function that can distinguish between classes, which is committed to improving the generalization ability and accuracy of our age estimation network.

\section{Experiment}
\label{sec:sec4}
In this section, we introduce the datasets and follow the evaluation criteria utilized in the experimental procedures. In addition, we present the specifics of our experiment and compare the results with state-of-the-art methods to validate the efficacy of our proposed technique. Finally, an ablation analysis is performed on the key elements within our approach to clarify how various aspects influence the overall performance.

\subsection{Datasets}
\textcolor{black}{\textbf{MORPH-II:}}
The dataset \citep{ricanek2006morph} is a large collection of cross-age facial images widely used in facial analysis studies. The dataset encompasses 55,134 facial photos featuring 13,000 distinct individuals, spanning an extensive age spectrum from 16 to 77 years. This experiment references two widely used evaluation schemes: \textbf{Setting I}: following \citep{gao2018age}, the dataset is randomly partitioned into two distinct segments, with the training set comprising 80$\%$ and the test set accounting for 20$\%$. \textbf{Setting II}: following \citep{ tan2017efficient}, we select a subset of 5493 facial images from Caucasian ethnic groups, which is split into two separate portions, allocating 80$\%$ for training and 20$\%$ for testing.

\textcolor{black}{\textbf{FG-NET:}}
The dataset \citep{lanitis2002toward} encompasses 1002 facial photographs, representing 82 individuals with ages varying from 0 to 69 years, demonstrating a range of ages from children to the elderly. The dataset provides information on 68 key points of the face in each image, which facilitates when performing tasks such as facial aging simulation and facial expression analysis. We followed the setup of previous methods \citep{ pan2018mean} using leave-one-person-out (LOPO) cross-validation.

\textcolor{black}{\textbf{ChaLearn LAP 2016:}}
The dataset \citep{escalera2016chalearn} is used in a challenge contest organized by ChaLearn focusing on the estimation of the appearance age of faces. It contains normally distributed age labels based on the annotation results of at least 10 people, which is very due for age estimation tasks that need to deal with real-world conditions. This dataset can be divided into a training set of 4,113 images, a validation set of 1,500 images, and a test subset of 1978 images.

\textcolor{black}{\textbf{UTK-Face:}}
The dataset \citep{zhang2017age} encompasses a vast age range that spans from infancy at 0 years to elderly at 116 years. This extensive collection, comprising more than 20,000 photographs, is annotated with demographic information such as gender, age, and ethnicity, thereby encapsulating a diverse array of human features. In our work,  we employ a stratified sampling approach, allocating 80$\%$ of the dataset for training purposes and reserving the remaining 20$\%$ for the testing of our models.

\subsection{Evaluation Criteria}
\subsubsection{\textbf{MAE(Mean Absolute Error)}}
It refers to the mean absolute error between the estimated age and the actual label value, which can be expressed as follows:
\begin{equation}
		MAE=\frac{1}{N}\sum_{i=1}^N{|}y_i-\hat{y}_i|
	\label{equ13}
\end{equation}
where $y_i$ and $\hat{y}_i$ denote the actual and estimated age values of the $i$-th sample, respectively, and $N$ is the amount of test images. The lower the value of the $MAE$, the superior the model's prediction performance.

\subsubsection{\textbf{$\epsilon$-error (Normal Score)}}
In the LAP-2016 dataset, the age is indicated as an average derived from various individuals' labels, and the true age in the data contains both mean and variance attributes. Therefore considering these factors can be a more accurate measure of the age estimation performance. A smaller $\epsilon$-error indicates improved performance in age estimation, and it can be expressed as follows:

\begin{equation}
		\epsilon =1-\sum_{i=1}^N{e}xp\left( -\frac{\left( y_i-\hat{y}_i \right) ^2}{2\sigma _i^2} \right)
	\label{equ14}
\end{equation}
where $y_i$ and $\hat{y}_i$ are the actual and estimated age values of the $i$-th sample, respectively, $N$ is the amount of test images, and $\sigma _i^2$ is the labeled standard deviation.

\subsubsection{\textbf{CS(Cumulative Score)}}
It focuses on the accuracy of the prediction within a certain error, which can be calculated as:
\begin{equation}
		CS\left( j \right) =\frac{N_{e\leq j}}{N}\times 100\%
	\label{equ15}
\end{equation}
where $N_{e\le j}$ indicates the number of image tests where the absolute error of the estimate does not exceed $j$. $N$ denotes the number of test images. The larger its value, the method is more robust.

\subsection{Implementation Details}
First, we utilize MTCNN \citep{zhang2016joint} for full face detection, align the image using the identified facial landmarks, and then crop and adjust it to a size of 224 * 224 pixels. Throughout the training phase, the image undergoes augmentation through random inversion, rotation, and translation. For the entirety of our experiments, we applied the Adam optimizer \citep{kingma2014adam}, with decay and momentum parameters set at 0.0005 and 0.9, respectively. The learning rate was initialized at 0.001, with a decay strategy based on cosine annealing. Utilizing PyTorch on an NVIDIA RTX 3090 GPU, we trained our framework over 120 epochs with a batch size of 32.

\textcolor{black}{
Meanwhile, during the experimental tuning process, considering the model performance and parameter overhead, we set the number of LRA-GNN layers $L$ to 12, the number of multi-head $M$ to 8, the loss function hyperparameter $\eta$ to 0.5, the initial graph construction threshold to 0.936, and random walk updating threshold to 0.824.
}

\subsection{Comparisons With The State-of-the-Art Methods}
To demonstrate the efficacy of our LRA-GNN, we perform comprehensive experiments across three facial image datasets. For the characteristics of different datasets, we utilize appropriate experimental settings and evaluation criteria to compare them with the state-of-the-art (SOTA) methods.

\subsubsection{\textbf{Comparisons on Morph II}}
On the most widely used Morph II dataset, we list the SOTA works year by year and give the backbone network used for that work as well as the number of parameters. We follow different protocols for comparison from the point of view of MAEs and the number of parameters. 
\textcolor{black}{
Table \ref{table1} shows the detailed results, where our method achieves an excellent performance of 1.79 (Setting I) and 1.94 (Setting II) with pre-training on an external dataset. Under Setting I, we only underperform HR \citep{hiba2023hierarchical}, GLAE\citep{bao2023general} and TAA-GCN\citep{korban2023taa}. HR proposed a hierarchical model integrating discrete age predictions with a collection of specialized regressors, each fine-tuned to enhance the probability estimation within specific age brackets. 
GLAE proposed Feature Rearrangement (FR) and Pixel-level Auxiliary learning (PA) to
take advantage of the facial features.
TAA-GCN designed the Temporal Memory Module (TMM) and Adaptive Graph Convolutional Layer (AGCL) to model age aging and temporal dependence.
}
Under Setting II, we achieved the best results. Thanks to our comprehensive capturing of latent relations, our LRA-GNN achieves performance similar to or even surpassing the SOTAs. At the same time, while achieving notable performance, we have fewer network parameters, which is attributed to the effectiveness and low redundancy of GCN in modeling facial key points.

\begin{table*}[!t]
	\caption{Comparison with the state-of-the-art methods on Morph II dataset. (*indicates used the extra datasets for pre-training.)}
	\renewcommand\arraystretch{1.4}
	\tabcolsep=0.025\linewidth
	\centering
	\begin{tabular}{c|ccccc}
		\hline
		\multirow{2}{*}{\textbf{Method}} &
		\multirow{2}{*}{\textbf{Venue Year}} &
		\multirow{2}{*}{\textbf{Backbone Network}} &
		\multicolumn{2}{c}{\textbf{MAE}} &
		\multirow{2}{*}{\textbf{Param.}} \\ \cline{4-5}
		&            &                  & \textbf{Setting I} & \textbf{Setting II} &       \\ \hline
		DEX\citep{rothe2015dex}          & IJCV 2016  & VGG-16           & -                  & 3.15/2.68*          & 138M  \\
		Ranking-CNN\citep{chen2017using} & CVPR 2017  & Binary CNNs      & 2.96*              & -                   & 500M  \\
		DLDLF\citep{shen2017label}       & NIPS 2017  & VGG-16           & 2.24               & -                   & 138M  \\
		MV\citep{pan2018mean}            & CVPR 2018  & VGG-16           & 2.79 /2.16*        & -                   & 138M  \\
		SSR-NET\citep{yang2018ssr}       & IJCAI 2018 & SSR-NET          & 3.16*              & -                   & 40.9K \\
		C3AE\citep{zhang2019c3ae}        & CVPR 2019  & C3AE             & 2.78*              & 2.95*               & \textbf{39.7K} \\
		BrigeNet\citep{li2019bridgenet}  & CVPR 2019  & VGG-16           & 2.38*              & 2.35*               & 138M  \\
		PML\citep{deng2021pml}           & CVPR 2021  & ResNet-34        & 2.15               & 2.31                & 21M   \\
		MWR\citep{shin2022moving}        & CVPR 2022  & VGG16            & 2.00              & 2.13               & 138M  \\
		MetaAge\citep{li2022metaage}        & TIP 2022  & VGG16            & 1.81              & 2.23               & 138M  \\
		DAA\citep{chen2023daa}           & CVPR 2023  & ResNet-18        & 2.25/2.06*         & -                   & 11M   \\
		TAA-GCN\citep{korban2023taa}           & PR 2023  & TAA-GCN        & 1.69         & -                  & -   \\
		DCT\citep{bao2022divergence}           & TIFS 2023  & ResNet-50        & 2.28/2.17*         & -                  & 23M   \\
		MSL\citep{wang2023exploiting}           & TIFS 2023  & ResNet-34        & 2.10         & 2.03                 & 21M   \\
		HR\citep{hiba2023hierarchical}           & TPAMI 2023  & VGG16        & \textbf{1.13*}         & 2.53*                 & 138M   \\
		
		\color{black}GLAE\citep{bao2023general}           & 	\color{black}TIP 2023  & 	\color{black}ResNet-50        & 	\color{black}1.14*         & 	\color{black}2.00*                 & 	\color{black}23M   \\
		
		\color{black}GroupFace\citep{zhang2024groupface}           & 	\color{black}TIFS 2024  & 	\color{black}EMGCN        & 	\color{black}2.09/1.86*         & 	\color{black}2.27/2.01*                 & 	\color{black}8.6M   \\
		
		\color{black}\citet{zhao2024mixture}           & 	\color{black}ISCI 2024  & 	\color{black}ResNet34(1/4)        & 	\color{black}1.85*         & 	\color{black}2.42*                 & 	\color{black}-   \\

		\cellcolor{gray!15}\textbf{LRA-GNN(Ours)} &
		\cellcolor{gray!15}\textbf{-} &
		\cellcolor{gray!15}\textbf{LRA-GNN} &
		\cellcolor{gray!15}2.02/1.79* &
		\cellcolor{gray!15}2.21/\textbf{1.94*} &
		\cellcolor{gray!15}13M \\ \hline
	\end{tabular}
	\label{table1}	
\end{table*}

\subsubsection{\textbf{Comparisons on FG-NET}}
We use the Mean Absolute Error (MAE) and Cumulative Score (CS) metrics on this few-shot dataset to compare with the SOTAs. CS represents the proportion of images where the absolute error does not exceed a threshold of $j$, following previous work with a setting of $j = 5$.
\textcolor{black}{
 As shown in Table \ref{table2}, our work earns the lowest MAE of 2.14 and the highest CS of 91.6$\%$, better than the similar GNN-based method MCGRL\citep{shou2025masked} that not capture the latent relations. 
}
This may be due to the fact that we have designed a series of graph enhancement strategies such as random walk and latent relation capturing, which make our model capable of learning discriminative features even in the face of fewer datasets.

\begin{table}[!t]
	\caption{Comparison with the state-of-the-art methods on FG-NET dataset. (*indicates used the extra datasets for pre-training.)}
	\renewcommand\arraystretch{1.4}
	\tabcolsep=0.016\linewidth
	\centering
	\begin{tabular}{c|ccc}
		\hline
		\textbf{Method}                                 & \textbf{MAE}  & \textbf{CS (\%)} & \textbf{Param.} \\ \hline
		DEX\citep{rothe2015dex}         & 4.63/3.09*    & 72.4             & 138M            \\
		DRFs\citep{shen2018deep}       & 3.85          & 80.6             & 138M            \\
		MV\citep{pan2018mean}           & 4.10/2.68*    & -                & 138M            \\
		C3AE\citep{zhang2019c3ae}       & 2.95*         & -                & \textbf{40.9K}  \\
		BrigeNet\citep{li2019bridgenet} & 2.56          & 86.0             & 138M            \\
		PML\citep{deng2021pml}          & 2.16*         & -                & 16M             \\
		MWR\citep{shin2022moving}       & 2.23 & 91.1             & -               \\
		DAA\citep{chen2023daa}          & 2.19*         & -                & 11M             \\
		TAA-GCN\citep{korban2023taa}           & 3.58        & -                  & -   \\
		\color{black} MCGRL\citep{shou2025masked}           & \color{black}2.86        & \color{black}88.0                 & \color{black}-   \\
		
		\cellcolor{gray!15}\textbf{LRA-GNN(Ours)}                          & \cellcolor{gray!15}\textbf{2.14*} & \cellcolor{gray!15} \textbf{91.6}    & \cellcolor{gray!15}13M             \\ \hline
	\end{tabular}
	\label{table2}	
\end{table}

\subsubsection{\textbf{Comparisons on ChaLearn LAP 2016}}
To enhance the assessment of our approach's efficacy in unconstrained conditions, we compared it against SOTAs on the ChaLearn LAP 2016 dataset. CLAP2016 is a challenging dataset done by multiple annotators, with the presence of manually subjectively labeled mean and variance, and thus we evaluate the model performance in conjunction with the $\epsilon$-error. Table \ref{table3} demonstrates the final results, where our method attains the minimal MAE at 3.11 and the minimal $\epsilon$-error at 0.258 for a moderate number of parameters. Compared to past models based on CNN architectures, our LRA-GNN achieves a more comprehensive feature extraction in a more flexible way, demonstrating the superiority of our method.

\begin{table}[!t]
	\caption{Comparison with the state-of-the-art methods on ChaLearn LAP 2016 dataset. (*indicates used the extra datasets for pre-training.)}
	\renewcommand\arraystretch{1.4}
	\tabcolsep=0.034\linewidth
	\centering
	\begin{tabular}{c|ccc}
		\hline
		\textbf{Method}             & \textbf{MAE}  & \textbf{$\epsilon$-error} & \textbf{Param.} \\ \hline
		AGEn\citep{tan2017efficient} & 3.82          & 0.3100          & 138M            \\
		MV\citep{pan2018mean}        & 3.14          & 0.287           & 138M            \\
		DLDL-v2\citep{gao2018age}    & 3.45          & 0.267           & \textbf{3.7M}   \\
		RAGN \citep{duan2018hybrid}  & -             & 0.367           & -               \\
		DCDL\citep{sun2021deep}      & 3.33          & -               & 138M            \\
		MetaAge\citep{li2022metaage} & 3.49          & 0.265           & 138M            \\
		MSL\citep{wang2023exploiting}
		& 3.25          & -           & 21M            \\
		\cellcolor{gray!15}\textbf{LRA-GNN(Ours)}      & \cellcolor{gray!15}\textbf{3.11*} & \cellcolor{gray!15}\textbf{0.258}  & \cellcolor{gray!15}13M             \\ \hline
	\end{tabular}
	\label{table3}
\end{table}

\subsubsection{\textbf{Comparisons on UTK-Face}}
We evaluated the performance of LRA-GNN on a large-scale, unconstrained dataset spanning ages from 0 to 116 years. As seen in Table \ref{table4}, GroupFace achieved the lowest Mean Absolute Error (MAE) of 4.22, significantly outperforming previous methods with a relatively modest number of parameters, totaling 13M. It is worth noting that	MSL\citep{wang2023exploiting} employed a deeper ResNet-34, and MWR\citep{shin2022moving} utilized a larger VGG16 architecture, both effectively reducing the MAE. Conversely, Andrey Savchenko \citep{savchenko2019efficient} utilized the more compact MobileNet-v2 with the fewest parameters but lagged in performance. These comparisons collectively demonstrate the effectiveness and reliability of our network across various types of facial datasets.

\begin{table}[!t]
	\caption{Comparison with the state-of-the-art methods on UTK-Face dataset. (* indicates used the extra datasets for pre-training.)}
	\renewcommand\arraystretch{1.4}
	\tabcolsep=0.012\linewidth
	\centering
	\begin{tabular}{c|ccc}
		\hline
		\textbf{Method}                                       & \textbf{Backbone} & \textbf{MAE}  & \textbf{Param.} \\ \hline
		\citet{savchenko2019efficient}       & MobileNet-v2      & 5.44*          & \textbf{3.4M}   \\
		CORAL \citep{cao2020rank}                          & ResNet-50         & 5.47*          & 25.6M           \\
		DCDL\citep{sun2021deep}                              & VGG16             & 4.48*          & 138M            \\
		MWR\citep{shin2022moving}                             & VGG16             & 4.37*          & 138M            \\
		MSL \citep{wang2023exploiting}           & ResNet-34        & 4.31                 & 21M   \\
		MIVOLO\citep{kuprashevich2023mivolo} & VOLO-D1       & 4.23*                 & 25.8M   \\
		\cellcolor{gray!15}\textbf{LRA-GNN(Ours)}                              & \cellcolor{gray!15}\textbf{LRA-GNN}    & \cellcolor{gray!15} \textbf{4.22}& \cellcolor{gray!15}13M            \\ \hline
	\end{tabular}
	\label{table4}
\end{table}

\subsection{Ablation Studies}
To demonstrate the efficacy of our method and strategy, we split the different components for a series of ablation experiments on several datasets without loading the eternal dataset pre-trained weights. We divide our method into three main constituent components: Latent Relation Capturing (LRC), Deep Feature Extraction (DFE), and Progressive RL-based Age Estimation (PRLAE). We first explore the effectiveness of these three components as a whole, and then further conduct more detailed ablation experiments on each of the three components to demonstrate the performance improvement of different modules.

\subsubsection{\textbf{Impact of Different Components}}
To test the impact of different components for age estimation accuracy, we set baseline as common DeepGCN, and then add components one by one for the experiment. As shown in Table \ref{table5}, the results give the MAEs of common GCN, and GCN with capturing latent relations, partial LRA-GNN, and full LRA-GNN on the three datasets. It can be seen that Latent Relation Capturing (LRC) contributes the most to enhancing the accuracy of age estimation, reducing the MAE by 0.28 in CLAP 2016. This may be due to the fact that unrestricted face samples are more complex and more necessary to capture the latent relations of facial key points. Secondly, Progressive RL-based Age Estimation (PRLAE) is also critical to performance improvement, directly reducing the MAE by 0.25 under Setting II in Morph II. 
\textcolor{black}{
Thirdly, Deep Feature Extraction (DFE) provides a significant gain to the model as well, directly reducing the MAE by 0.29 in the FG-NET dataset, while the combination of DFE and LRC enables a more comprehensive and deeper facial representation.
} 

\begin{table*}[!t]
	\caption{The impact of different key components.}
	\renewcommand\arraystretch{1.4}
	\setlength\tabcolsep{5.5mm}
	\centering
	\begin{tabular}{ccc|ccccc}
		\hline
		\multicolumn{3}{c|}{Components} & \multicolumn{2}{c}{Morph II} & \multirow{2}{*}{FG-NET} & \multirow{2}{*}{CLAP 2016} & \multirow{2}{*}{UTK-Face} \\ \cline{1-5}
		LRC      & DFE      & PRLAE     & Setting I    & Setting II    &                         &                            &                           \\ \hline
		
	    -	&             - &               - & 2.47               & 2.88                & 2.68                             & 3.54           & 4.62                     \\
		
		\textbf{\checkmark}   &   -           &         -       & 2.29               & 2.64                & 2.41                             & 3.26          & 4.38                      \\
		
		-   &   \color{black}\textbf{\checkmark}          &       \color{black}  -       & \color{black} 2.35               & \color{black} 2.73                & \color{black}2.39                             & \color{black}3.32          & \color{black}4.45                      \\

	    \textbf{\checkmark}   & \textbf{\checkmark}   &    -            & 2.18               & 2.46                & 2.32                             & 3.21 & 4.34                                \\
	
	   - &\textbf{\checkmark}    &  \textbf{\checkmark}               & 2.21               & 2.43                & 2.28                             & 3.24 & 4.41                                \\
	
		\textbf{\checkmark}   &  -  & \textbf{\checkmark}               & 2.15               & 2.39                & 2.24                             & 3.18 & 4.28                                \\
	
		\cellcolor{gray!15} \textbf{\checkmark}   & \cellcolor{gray!15} \textbf{\checkmark}   & \cellcolor{gray!15} \textbf{\checkmark}     & \cellcolor{gray!15} \textbf{2.02}      & \cellcolor{gray!15} \textbf{2.21}       & \cellcolor{gray!15} \textbf{2.14}                    & \cellcolor{gray!15} \textbf{3.11}             &      \cellcolor{gray!15} \textbf{4.22}    \\ \hline
	\end{tabular}
	\label{table5}
\end{table*}

\begin{table*}[!t]
	\caption{The impact of different types of Random Walk Updating.}
	\renewcommand\arraystretch{1.4}
	\setlength\tabcolsep{7.8mm}
	\centering
	\begin{tabular}{c|ccccc}
		\hline
		\multirow{2}{*}{Types} & \multicolumn{2}{c}{Morph II} & \multirow{2}{*}{FG-NET} & \multirow{2}{*}{CLAP 2016} & \multirow{2}{*}{UTK-Face} \\ \cline{2-3}
		& Setting I    & Setting II    &                         &                            &                           \\ \hline

		\textbf{BFS}                & 2.13               & 2.32                & 2.28                             & 3.23       & 4.31                                                                    \\
		
		\textbf{DFS}         & 2.07               & 2.26                & 2.21                             & 3.19         & 4.28                                                                 \\
		
		\cellcolor{gray!15}\textbf{BFS + DFS}  & \cellcolor{gray!15}\textbf{2.02}      & \cellcolor{gray!15}\textbf{2.21}       & \cellcolor{gray!15}\textbf{2.14}                    & \cellcolor{gray!15}\textbf{3.11}
		&\cellcolor{gray!15}\textbf{4.22}                                                     \\ \hline
	\end{tabular}
	\label{table6}
\end{table*}

\begin{table*}[!t]
	\caption{The impact of progressive RL-based age estimation.}
	\renewcommand\arraystretch{1.4}
	\setlength\tabcolsep{6.6mm}
	\centering
    \begin{tabular}{c|ccccc}
	\hline
	\multirow{2}{*}{Designs} & \multicolumn{2}{c}{Morph II} & \multirow{2}{*}{FG-NET} & \multirow{2}{*}{CLAP 2016} & \multirow{2}{*}{UTK-Face} \\ \cline{2-3}
	& Setting I    & Setting II    &                         &                            &                           \\ \hline

		\textbf{Decouple}                & 2.18               & 2.46                & 2.32                             & 3.21       & 4.34                                                                    \\
		
		\textbf{Co-Optimization}         & 2.14               & 2.37                & 2.25                             & 3.16         & 4.31                                                                 \\
		
		\textbf{+Imbalance Ratio}        & 2.09               & 2.28                & 2.19                             & 3.13       & 4.25                                                                    \\
		
		\cellcolor{gray!15}\textbf{+Distribution Distance}  & \cellcolor{gray!15}\textbf{2.02}      & \cellcolor{gray!15}\textbf{2.21}       & \cellcolor{gray!15}\textbf{2.14}                    & \cellcolor{gray!15}\textbf{3.11}
	  &\cellcolor{gray!15}\textbf{4.22}                                                     \\ \hline
	\end{tabular}
	\label{table7}
	\end{table*}

\subsubsection{\textbf{Impact of Latent Relation Capturing}}
In Latent Relation Capturing, we utilize facial key points as prior knowledge (FK) and Random Walk Strategy (RW) to jointly guide the effective capturing and comprehensive representation of latent relations. To analyze the gain of these two strategies on latent relation capturing and to discuss the impact of the quantity of $M$ heads in the mechanism of multi-head attention, we conduct experiments on the FG-NET dataset, which is evaluated using the cumulative score (CS).

\begin{figure}[!t]
	\centering
	\includegraphics[width=0.5\textwidth]{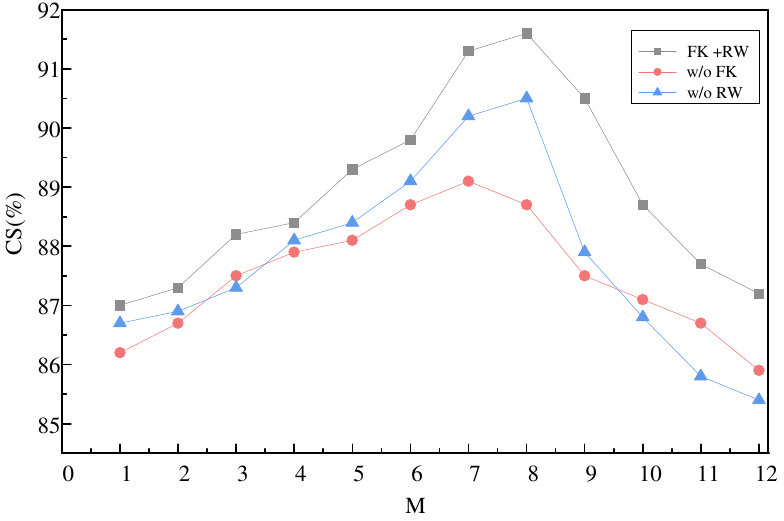}
	\caption{The ablation study of Latent Relation Capturing on the FG-NET dataset.}
	\label{fig_6}
\end{figure}

\textcolor{black}{
As shown in Figure \ref{fig_6}, both the utilization of face key points as prior knowledge and the random walk strategy are effective in improving the performance of age estimation. Without the joint guidance, as the number of fully connected graphs $M$ increases, more redundant information is easily generated, making it difficult to effectively capture latent relations. We observe that the most performance of 91.6$\%$ is achieved when the quantity of multi-head is 8, so we choose $M$ to 8.
}

We also explore the effect of specific strategy types in random walk updating on our graph model. As shown in Table \ref{table6}, considering both the Breadth First Search (BFS) and Depth First Search (DFS) can effectively reduce the age prediction error. This is because BFS tends to explore local neighbors and is difficult to capture long-distance dependencies, while DFS tends to explore more deeply but lacks a description of the exact nature of these dependencies. LRA-GNN flexibly combines the two which can better capture global-local relations.

\begin{figure}[!t]
	\centering
	\includegraphics[width=0.5\textwidth]{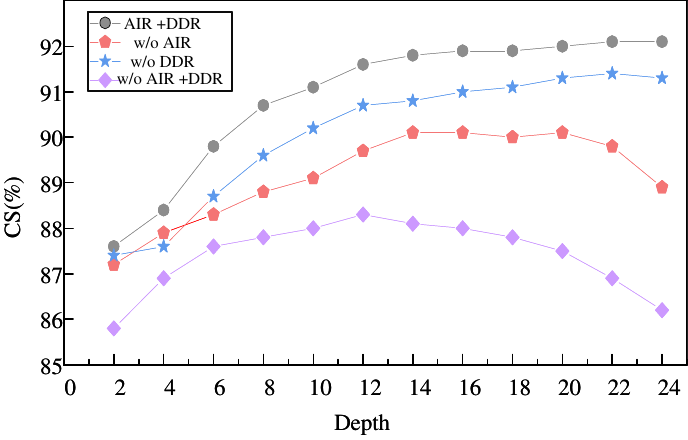}
	\caption{The ablation study of Deep Feature Extraction on the FG-NET dataset.}
	\label{fig_7}
\end{figure}

\subsubsection{\textbf{Impact of Deep Feature Extraction}}
We continue to evaluate the effectiveness of our Deep Feature Extraction. 
\textcolor{black}{
To quantify the contribution of the adaptive initial residual (AIR) and dynamic developmental residual (DDR) of our design, keeping the other components unchanged, four objects were set up: AIR+DDR, w/o AIR, w/o DDR and w/o AIR+DDR. As shown in Figure \ref{fig_7}, the performance without the adaptive initial residual (AIR) and dynamic developmental residual (DDR) grows with the number of layers starts to degrade after a shallower growth, which could be the introduction of noise leading to over-smoothing problems. With AIR or DDR mitigates the over-smoothing problem somewhat, where with AIR performance starts to decline after reaching the optimum, and the upper limit of the performance without DDR is difficult to break through and improves slowly. Our Deep Feature Extraction fusing AIR and DDR can better alleviate the GCN over-smoothing problem and ensure the consistency and diversity of the information, thus effectively improving the age estimation performance. 
Considering the balance between performance and parameter count, the performance improvement is less but the number of parameters increases more after $L>12$, so we set the number of layers to 12.
}

\subsubsection{\textbf{Impact of Progressive RL-based Age Estimation}}
In Progressive RL-based Age Estimation, we define age estimation by classification as a walk-to-the-end problem on a grid, where the grid is divided into rows and columns. As shown in Table \ref{table7}, our age estimation method allows for simultaneous optimization of classification and regression, achieving a considerable improvement over the decoupling method. Besides, our well-designed reward function not only designs the imbalance ratio to guide the agent to better perform the behavior in the unbalanced dataset but also adds the distance from label value to take into account the continuity of the age distribution, which both significantly improves the generalization of age estimation. Under Setting II of Morph II, we get the maximum gain of 0.25 MAE.

In addition, we evaluate the impact of the parameter $\eta $ in the loss function. As shown in Table \ref{table8}, Morph II (Setting I) and FG-NET datasets achieved the best results at $\eta=0.5 $ and  $\eta=0.4 $, respectively. This is probably because Focal Loss is a benefit for coping with hard samples and class distributions, and jointly optimizing the two loss functions for categorical regression with about the same weights can obtain excellent results.

\begin{table}[!t]
	\caption{The impact of the parameter $\eta $ in the age estimation loss function. }
	\renewcommand\arraystretch{1.4}
	\tabcolsep=0.014\linewidth
	\centering
 \begin{tabular}{c|lllllllll}
	\hline
	$\eta $                           & 0.1 & 0.2 & 0.3 & 0.4  & 0.5  & 0.6 & 0.7 & 0.8 & 0.9 \\ \hline
	\multicolumn{1}{l|}{Morph II} &  2.21   &   2.15  &  2.12   &  2.06    & \textbf{2.02} &   2.05  &  2.13   &  2.18   &  2.25   \\
	FG-NET                        &  2.29   &  2.25   &  2.21   & \textbf{2.14} &   2.16   &  2.23   &  2.27   &  2.31   &  2.33   \\ \hline
  \end{tabular}
	\label{table8}
\end{table}

\begin{figure*}[!t]
	\centering
	\includegraphics[width=1\textwidth]{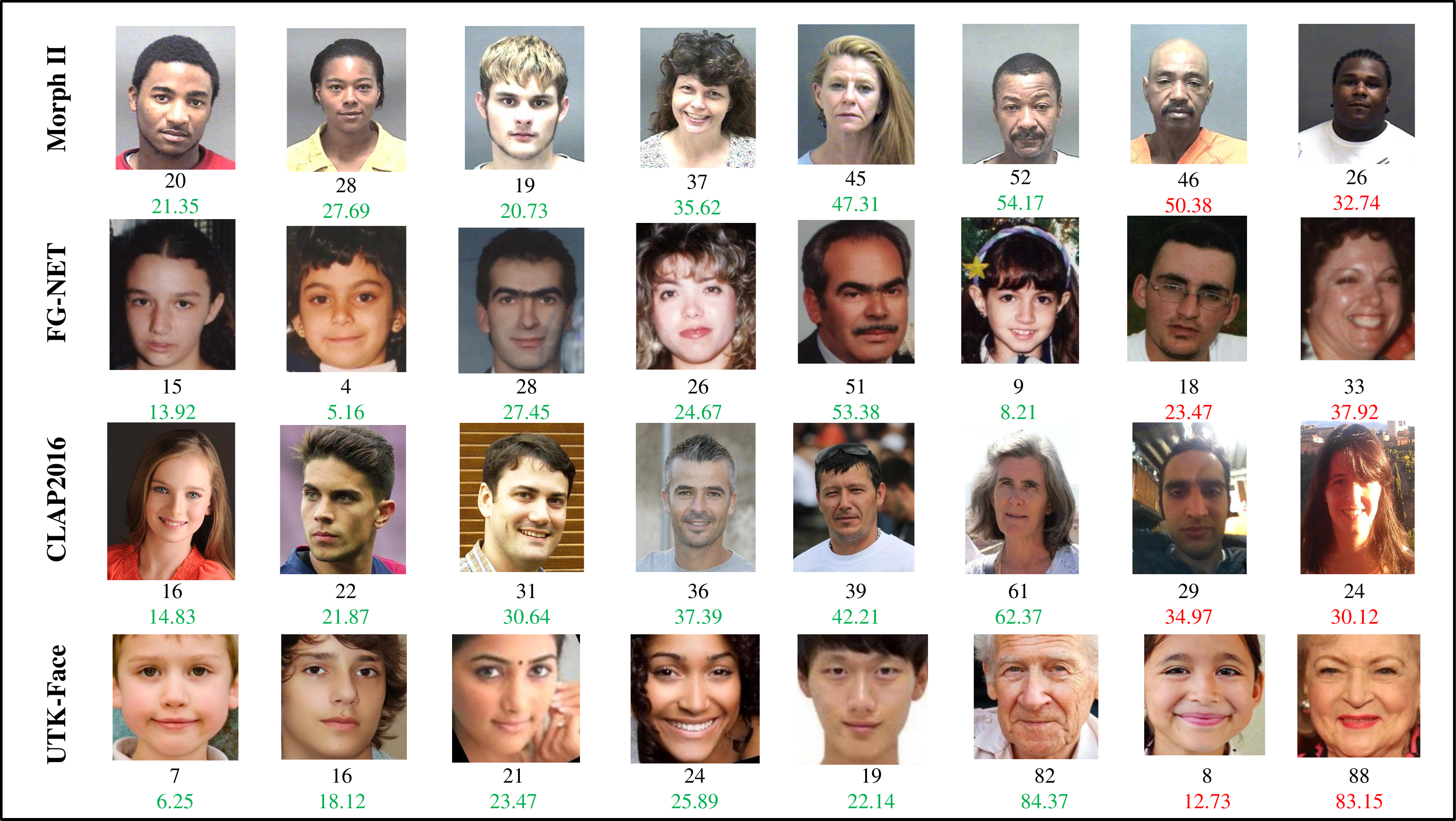}
	\caption{The examples of age estimation results of our LRA-GNN on three facial datasets. The factual label is the black number, the reliable estimation results are shown in the green number, and the poor estimation results are shown in the red number.}
	\label{fig_9}
\end{figure*}

\subsection{Qualitative Results}
To better qualitative the effectiveness of our method, we first discuss the prediction accuracy of LRA-GNN for different age groups and analyze the possible reasons for the results. Then, we measure and compare the runtime and number of parameters of our architecture to analyze the efficiency. Finally, we randomly select some samples in the three datasets to demonstrate the prediction results and analyze the reasons for the success and failure of age estimation.

\begin{figure}[!t]
	\centering
	\subfloat[On Morph II]{\includegraphics[width=0.48\textwidth]{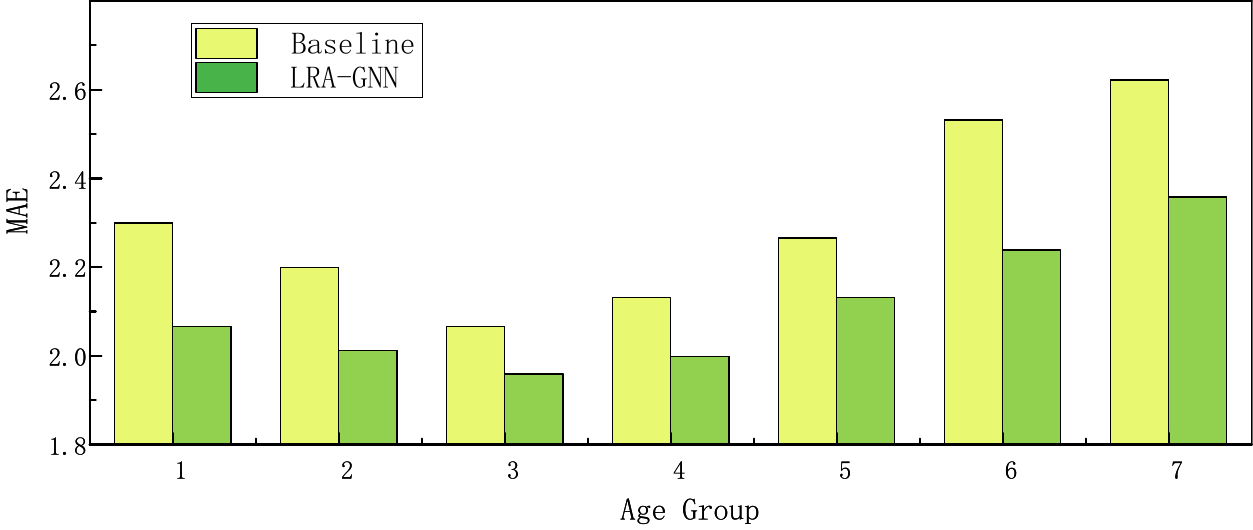}%
		\label{On Morph II}}
	\hfil
	\subfloat[On FG-NET]{\includegraphics[width=0.48\textwidth]{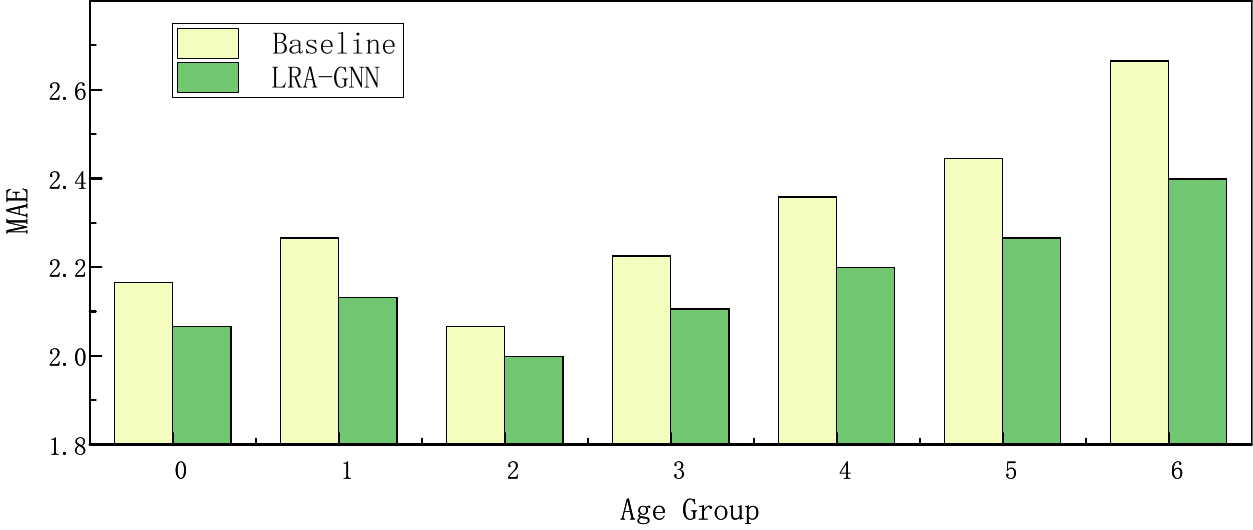}%
		\label{On FG-NET}}
	\caption{The accuracy analysis of the baseline and our LRA-GNN on Morph II and FG-NET datasets.}
	\label{fig_8}
\end{figure}

\subsubsection{Accuracy Analysis}
We discuss the estimation accuracy of LRA-GNN on Morph II (Setting I) and FG-NET datasets for different age groups. Baseline utilizes a common DeepGCN without capturing latent relations. As shown in Figure \ref{fig_8}, the performance of each age group of our LRA-GNN on both datasets is somewhat improved compared to the Baseline. The most obvious improvement is in the $g_1$ (10-19 years old) age group of Morph II and $g_6$ (60-69 years old) of FG-NET, probably due to our mining of latent relations and the design of reward function improving the hard sample representation more.

\begin{figure}[!t]
	\centering
	\includegraphics[width=0.48\textwidth]{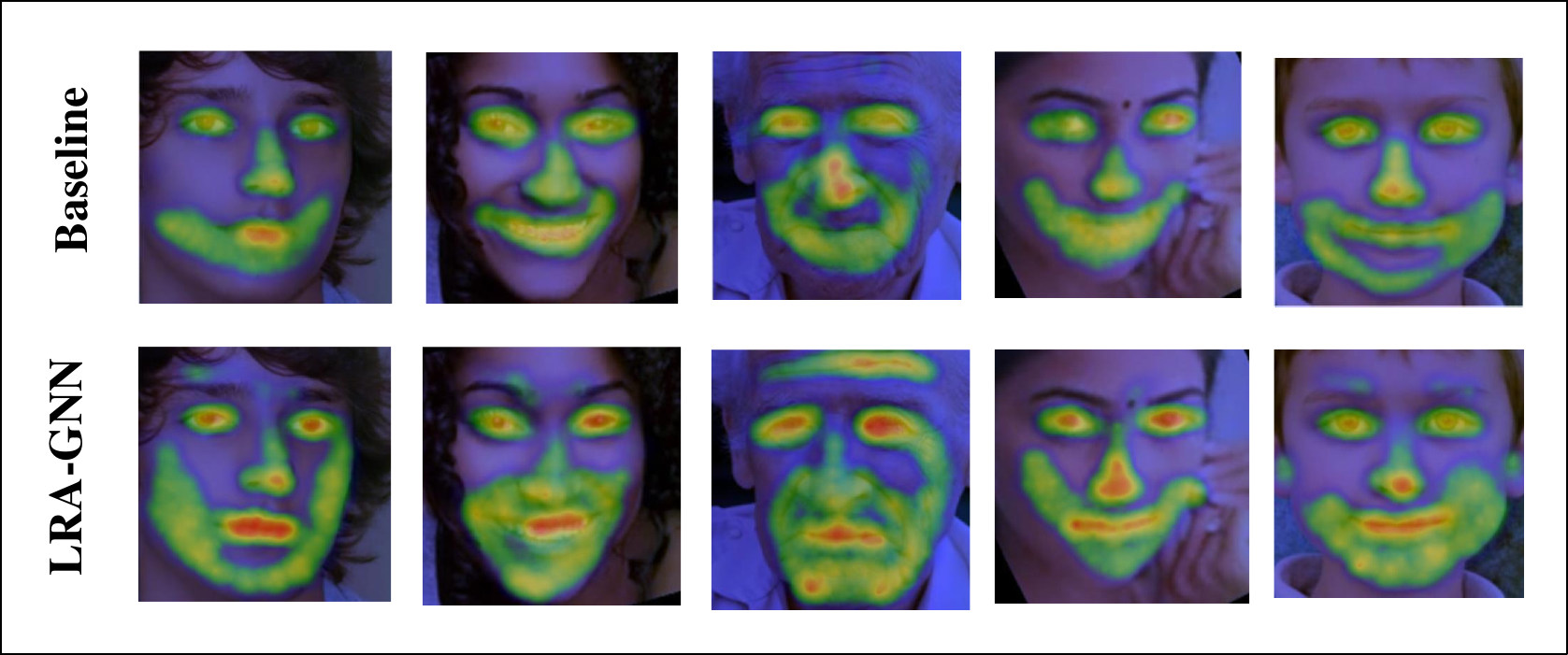}
	\caption{\textcolor{black}{The visualization examples of LRA-GNN for facial feature extraction utilizing attention heat map.}}
	\label{fig_10}
\end{figure}

\subsubsection{Efficiency Analysis}
\textcolor{black}{
We first perform a theoretical analysis of the time complexity of the main design part of the LRA-GNN. For the construction of the graph, we have $N$ nodes (i.e., facial keypoints) with $m$ features per node, which mainly depends on calculating the similarity between nodes, which takes $\mathcal{O}\left( N^2\times m \right) $ time. The time complexity of a random walk depends on the number of steps $t$ of the walk and the sparsity of the graph. For each node, we need to update the probability of its neighboring nodes, which takes $\mathcal{O}\left( t\times d \right)$ time, where $d$ is the average degree of the graph. The time complexity of the multi-head attention mechanism mainly depends on the number of heads $M$ and the amount of computation for each head. For each head, we need to compute the attention weights between $N$ nodes, which takes $\mathcal{O}\left( N^2\times M \times m \right)$ time. The time complexity of the deep residual graph convolutional network depends on the number of layers $L$ and the amount of computation per layer. For each layer, we need to perform graph convolution operation which takes $\mathcal{O}\left( N^2\times L \times m \right)$ time. The time complexity of the reinforcement learning part depends on the size of the state space $S$, the size of the action space $A$ and the number of training iterations $T$, which is $\mathcal{O}\left( S\times A \times T \right)$. Thus the total time complexity can be expressed as $\mathcal{O}\left( \left( M+L \right) N^2\times m+t\times N\times d+S\times A\times T \right) $.
}

\textcolor{black}{
Then we perform the efficiency comparison tests using GPUs under the same experimental setup on UTK-Face dataset.
}
We compare the reasoning runtime of CNN \citep{shin2022moving, wang2023exploiting}, Transformer \citep{kuprashevich2023mivolo} and GNN \citep{zhang2024multi}. As shown in Table. \ref{table9}, compared to ViG-S, which does not capture latent relations, LRA-GNN increases on the fold of the computational overhead but achieves the lowest MAE and number of parameters. Moreover, the running time of LRA-GNN still outperforms VGG-16 and VOLO-D1, which shows that our architecture is worthwhile and reliable.

\begin{table}[!t]
	\caption{The efficiency analysis on UTK-Face dataset. }
	\renewcommand\arraystretch{1.4}
	\tabcolsep=0.045\linewidth
	\centering
	\begin{tabular}{c|ccc}
		\hline
		\textbf{Architectures}                                       & \textbf{Runtime} & \textbf{Param}  & \textbf{MAE} \\ \hline
		
		VGG-16       & 149.29ms      & 138M         & 4.37   \\
		
		ResNet-34                            & \textbf{61.88ms}         & 21M          & 4.31           \\
		
		VOLO-D1                              & 194.74ms              &          25.8M & 4.23            \\
		
		ViG-S                           &    76.43ms          & 27.3M          & 4.28            \\
		
		\cellcolor{gray!15}\textbf{LRA-GNN(Ours)}                              & \cellcolor{gray!15}141.96ms    & \cellcolor{gray!15} \textbf{13M}& \cellcolor{gray!15}\textbf{4.22}            \\ \hline
	\end{tabular}
	\label{table9}
\end{table}

\subsubsection{Visualization Analysis}
We selected samples from three different types of facial datasets as a presentation of the estimation results. As shown in Figure \ref{fig_9}, our LRA-GNN achieves excellent performance on both constrained and unconstrained datasets. The green numbers show that our method performs well in different datasets and age groups, which can be due to the capturing of latent relations, as well as the progressive reinforcement learning-based co-optimization for age estimation. However, the red numbers show some of the poor estimation results, which may be caused by severe occluded faces, poor backgrounds, and so on.

\textcolor{black}{
Furthermore, we employ the attention heat map for visualization to compare  the face attention regions obtained from Baseline and LRA-GNN. As can be seen from Figure \ref{fig_10}, compared with Baseline, LRA-GNN covers a larger area of attention and provides better recognition of key points of the face such as eyes, mouth, and nose, and is more sensitive to the detection of some wrinkles. This proves that our network can capture more useful information for age estimation by utilizing the multi-head attention mechanism.
}

\section{Conclusion}
\label{sec:sec5}
In this novel, we have presented a new Latent Relation-Aware Graph Neural Network with Initial and Dynamic Residual (LRA-GNN) to achieve robust and comprehensive facial representation. We first construct an initial graph utilizing facial key points as prior knowledge, and then a random walk strategy is employed on the initial graph to obtain the global structure. The LRA-GNN leverages the multi-attention mechanism to capture the latent relations and generates a set of fully connected graphs. We also design the deep residual graph convolutional networks for deep feature extraction on the fully connected graphs, which fuse adaptive initial residuals and dynamic developmental residuals to ensure the consistency and diversity of information. Finally, we propose progressive reinforcement learning to co-optimize the ensemble classification regressor. Our proposed method outperforms the state-of-the-art methods on Morph II, FG-NET, and CLAP 2016 age estimation benchmarks, demonstrating its strength and effectiveness. In the future, we consider further optimization of the age estimation algorithm to achieve more robust results under unconstrained conditions.

\section*{Competing interests} The authors declare that they have no known competing financial interests or personal relationships that could have
appeared to influence the work reported in this paper.

\section*{Data availability and access} Data will be made available on request.
	
\section*{Acknowledgements}
	This work is supported by National Natural Science Foundation of China (Grant No. 69189338), Excellent Young Scholars of Hunan Province of China (Grant No. 22B0275).
	

\bibliographystyle{cas-model2-names}

\bibliography{refs.bib}


	
\end{document}